\documentclass[conference]{IEEEtran}
\usepackage{lipsum}
\usepackage{graphicx}
\usepackage{hyperref}
\usepackage{fancyhdr}
\usepackage[numbers,sort,compress]{natbib}
\hypersetup{hidelinks,
	colorlinks=true,
	allcolors=black,
	pdfstartview=Fit,
	breaklinks=true
}
\IEEEoverridecommandlockouts

\usepackage[T1]{fontenc}

\begin{document}
 \title{WF-UNet: Weather Fusion UNet for Precipitation Nowcasting}


\author{
    \IEEEauthorblockN{
        Christos Kaparakis\IEEEauthorrefmark{2}, Siamak Mehrkanoon \IEEEauthorrefmark{2}\IEEEauthorrefmark{3}\IEEEauthorrefmark{1}\thanks{*Corresponding author.} 
    }
    \IEEEauthorblockA{\IEEEauthorrefmark{2} 
    Department of Advanced Computing Sciences, Maastricht University, Maastricht, Netherlands}
    \IEEEauthorblockA{ \IEEEauthorrefmark{3} Department of Information and Computing Sciences, Utrecht University, Utrecht, Netherlands}
    christos.kaparakis@student.maastrichtuniversity.nl, 
    s.mehrkanoon@uu.nl
}



\maketitle
\renewcommand{\headrulewidth}{0pt}
\begin{abstract}
Designing early warning systems for harsh weather and its effects, such as urban flooding or landslides, requires accurate short-term forecasts (nowcasts) of precipitation. Nowcasting is a significant task with several environmental applications, such as agricultural management or increasing flight safety. In this study, we investigate the use of a UNet core-model and its extension for precipitation nowcasting in western Europe for up to 3 hours ahead. In particular, we propose the Weather Fusion UNet (WF-UNet) model, which utilizes the Core 3D-UNet model and integrates precipitation and wind speed variables as input in the learning process and analyze its influences on the precipitation target task. We have collected six years of precipitation and wind radar images from Jan 2016 to Dec 2021 of 14 European countries, with 1-hour temporal resolution and 31 square km spatial resolution based on the ERA5 dataset, provided by Copernicus, the European Union's Earth observation programme. We compare the proposed WF-UNet model to persistence model as well as other UNet based architectures that are trained only using precipitation radar input data. The obtained results show that WF-UNet outperforms the other examined best-performing architectures by 22\%, 8\% and 6\% lower MSE at a horizon of 1, 2 and 3 hours respectively.
\end{abstract}

\IEEEpeerreviewmaketitle


\section{Introduction}
\label{sec:intro}
Heavy rainstorms can impact people's daily lives negatively. In 2017, the WMO (World Meteorological Organization)
investigated the impact of precipitation extremes, in terms of both excess and deficient rainfall \footnote{\href{https://public.wmo.int/en/media/news/rainfall-extremes-cause-widespread-socio-economic-impacts}{https://public.wmo.int/en/media/news/rainfall-extremes-cause-widespread-socio-economic-impacts}}. Thousands of people died and were displaced due to flooding and landslides in many countries worldwide. These disruptions had widespread socio-economic impacts. Therefore, it would be advantageous to forecast these weather events in advance so that decision-makers can take steps to safeguard lives, properties and wealth.

Computational weather forecasting is an integral feature of modern, industrialized societies. It is used for planning, organization and managing many personal and economic aspects of life. Many industries such as agriculture \cite{bendre2015big}, mining \cite{bag2022role} and construction \cite{hazyuk2012optimal} rely on weather forecasts to make decisions, and if climatological events occur that are unexpected, this can lead to significant economic losses.  Similarly, accurate weather forecasts improve flight and road safety and help foresee potential natural disasters. In precipitation nowcasting, one attempts to provide a short-range forecast of rainfall intensity based on radar echo maps, rain gauges and other observation data \cite{lebedev2019precipitation}. Precipitation nowcasting is an important task because of its immense impact, for instance, in predicting road conditions \cite{kilpelainen2007effects}  and enhancing flight safety by providing weather guidance for regional aviation \cite{van2007operational}. Furthermore, it can help with flood mitigation, water resource management \cite{schwanenberg2015short} and avoiding casualties by issuing citywide rainfall alerts \cite{thielen2009european}.

Forecasting weather generally relies on Numerical Weather Prediction (NWP) \cite{bauer2015quiet}, methods based on mathematical formulations using several atmospheric features. Although NWP is a powerful method for such forecasting tasks, it also has drawbacks as it requires too much computational power \cite{al2010review}. In addition, forecasting with NWP may be sensitive to noise present in the measurements of weather variables \cite{ancell2018seeding, mehrkanoon2019deep}. NWP models may take hours to run and are also less accurate than persistence-based forecasts on less than 4 hour predictions \cite{dragoon2010valuing,ashok2022systematic}. In recent years, the enormous amount of ever-increasing weather data has stimulated research interest in data-driven machine learning techniques for nowcasting tasks \cite{agrawal2019machine,camporeale2019challenge, 9054232,FERNANDEZ2021419,chen2021transunet,trebing2021smaat}. Unlike the model-driven methods, data-driven models do not base their prediction on the calculations of the underlying physics of the atmosphere. Instead, they analyze and learn from historical weather data such as past wind speed and precipitation maps to predict the future. By taking advantage of available historical data, data-driven approaches have shown better performance than classical ones in many forecasting tasks \cite{dragoon2010valuing,ashok2022systematic}.

Recent advances in Artificial Neural Network architectures (ANNs) have demonstrated great potential in precipitation nowcasting tasks \cite{shi2015convolutional,wang2017predrnn,tran2019multi,sonderby2020metnet}. The critical difference between NWP and ANNs is that the former is model-driven, and the latter is a data-driven approach \cite{shi2017deep,trebing2021smaat, BILIONIS201422}. While classical machine learning techniques rely on handcrafted features and domain knowledge, deep learning techniques automatize the extraction of those features \cite{shaheen2016impact}. Deep learning techniques have demonstrated remarkable results in several domains such as biomedical signal analysis, healthcare, neuroscience and dynamical systems, among others \cite{aykas2021multistream,FERNANDEZ2021419,abdellaoui2021enhancing,zemouri2019deep, marblestone2016toward}. Due to underlying complex dynamics of weather data, ensuring accurate nowcasting at several temporal-spatial levels is a challenging task. Deep learning based models have recently received much attention in this area because of their powerful abilities to learn spatiotemporal feature representations in an end-to-end fashion \cite{trebing2021smaat, fernandez2021broad, yang2022aa,shi2015convolutional}. UNet, one of the successful deep learning architectures, which was originally proposed for image segmentation task, have shown promising results in various domains such as precipitation nowcasting, background detection in video analysis and image super-resolution \cite{trebing2021smaat,tao2017background,fernandez2021broad, tran2015learning, hu2019runet, yang2022aa}.
It consists of a contracting path to extract features, and an expanding path, to reconstruct a segmented image, with a set of residual connections between them to enable precise localization.

Most of the current state-of-the-art deep learning based models use only the past precipitation radar images to predict the future precipitations \cite{shi2015convolutional, shi2017deep, trebing2021smaat,fernandez2021broad}. However, classical weather forecasting which generally rely on Numerical Weather Prediction (NWP), uses several atmospheric features (weather variables). Generally, it is common for measurements from different modalities to carry complementary information about various aspects of a particular activity \cite{ngiam2011multimodal}. Therefore, data-driven models that combine data from multiple sources can potentially provide more accurate inferences. This research aims to develop a novel architecture that can incorporate multiple weather variables such as past precipitation together with wind speed, as input to more accurately forecast the precipitation.
This paper is organized as follows. A brief overview of the related research works is given in Section 2. Section 3 introduces the proposed WF-UNet model. The data description and used preprocessing steps are explained in Section 4. The experimental setup and and evaluation of the models are given in Section 5. The obtained results are discussed in Section 6 and the conclusion is drawn in Section 7.

\section{Related Work}
In recent years, data-driven models have gained much attention for predicting weather elements such as temperature, wind speed and precipitation \cite{fernandez2021broad,bilgin2021tent,aykas2021multistream,mehrkanoon2019deep}.
Due to the vast amount of available weather data and the fact that weather element forecasting can be formulated as a sequence prediction problem, deep neural networks architectures such as Recurrent Neural Network (RNN) \cite{9054232}, Long Short-Term Memory (LSTM) \cite{shi2015convolutional} and Convolutional Neural Network (CNN) \cite{rs13163278} among others are suitable candidates to address various problems in this field. 
In particular, CNN architectures have shown their excellent ability to handle 2D and 3D images. Moreover, thanks to the versatility of CNNs, nowcasting problems can be tackled in different fashions. For instance, the authors in \cite{ayzel2019all} and \cite{agrawal2019machine} treated the multiple timesteps as multiple channels in the network. This way, they could apply a 2D-CNN to perform the predictions. The authors in \cite{shi2017deep} also treated the multiple timesteps as depth in the samples and applied a 3D-CNN and approximate more complex functions. The authors in \cite{shi2015convolutional}, introduced an architecture that combines both CNN's and RNN's strengths to be able to more efficiently work with image data as well as capturing long-range dependencies for the precipitation nowcasting task.
Hybrid models like in \cite{he2016deep} replace the fully-connected LSTM with ConvLSTM to better capture the spatiotemporal correlations. In \cite{shi2015convolutional}, the network predicted raw pixels directly rather than predicting the transformation of the input frame whereas  \cite{finn2016unsupervised} predicted transformed pixels instead. 
In \cite{lebedev2019precipitation,trebing2021smaat,agrawal2019machine}, the authors show the successful application of UNet model in precipitation task due to its autoencoder-like architecture and ability to tackle image-to-image translation problems. 
\begin{figure*}[h!]
    \centering
    \includegraphics[width=18cm, height =8.5cm]{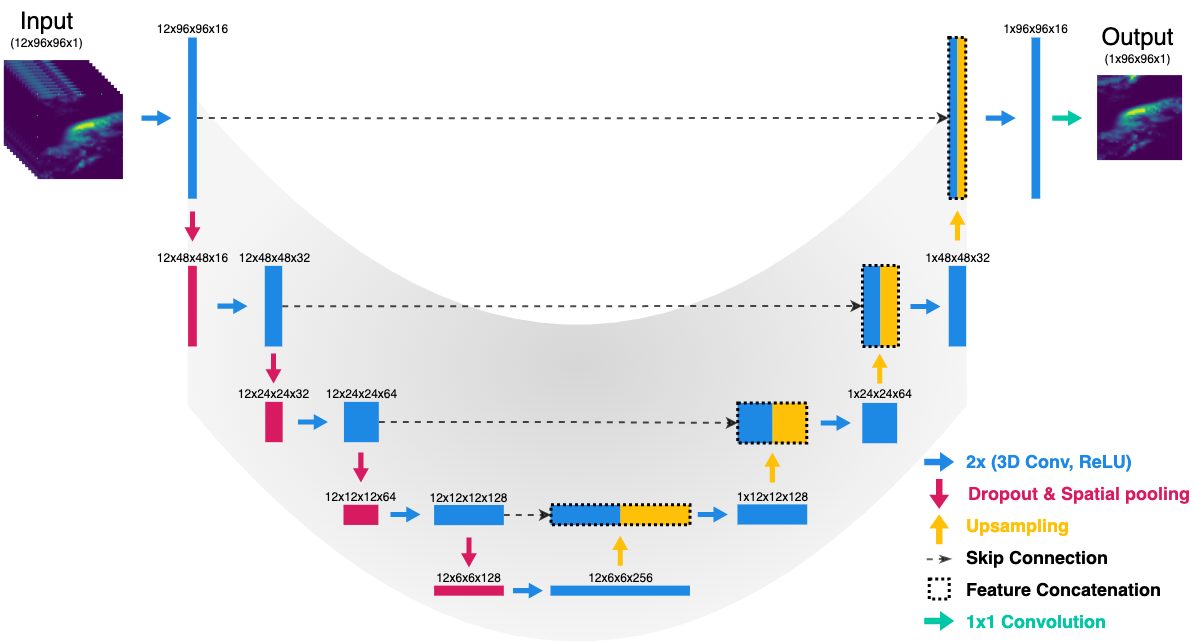}
    \caption{The Core UNet encoder-decoder network. A multi-channel feature map is represented by each cube in the figure. Numbers on the top-right of a cube indicate the number of channels, while numbers on the left of a cube indicate the resolution. Note that at each level of the architecture, the feature maps have the same resolution. The input is the ground truth of a sequence of precipitation maps, whereas the output is the corresponding prediction map produced by the model.}
    \label{fig:myu_net}
\end{figure*}
The SmaAt-UNet model, an extension of the UNet model, which significantly reduces the UNet parameters without compromising its performance, is introduced in \cite{trebing2021smaat}. Finally, the authors in \cite{FERNANDEZ2021419} introduced Broad-UNet by equipping the UNet model with asymmetric parallel convolutions and the Atrous Spatial Pyramid Pooling (ASPP) module. Despite its many advantages, UNet can still result in limited abilities when modelling long-range dependencies due to the intrinsic locality of the convolution-based operations. 

In \cite{chen2021transunet}, the researchers propose another variation of the UNet model, the TransUNet. CNN-Transformer is used as an encoder in TransUNet, whereas the original UNet decoder is used as a decoder. There has been success in applying this model to medical image segmentation tasks. Finally, in \cite{yang2022aa}, the researchers propose another variation of the UNet model, the AA-TransUNet. A pair of key elements are added to extend TransUNet: Convolutional Block Attention Modules (CBAM) and Depthwise-separable Convolutions (DSC). Therefore, the model can explore the inter-channel and inter-spatial relationships of features by performing both channel-wise and spatial-wise attention.

\section{Proposed Model}
This section introduces our proposed \textbf{WF-UNet} model which adopts a variation of the original UNet architecture as its core building block and combines it with a data fusion strategy in order to incorporate two different weather variables. 

\subsection{Core UNet model}
Compared to the original UNet \cite{ronneberger2015u}, here 3D Convolution layers are used. These convolutions can not only capture the spatial information contained within one radar image, but the spatial information of an image with its previous timesteps, i.e., the \textit{temporal} information. Fig. \ref{fig:myu_net} outlines the architecture of the used Core UNet model which consists of the encoder and an decoder paths. As shown in Fig. \ref{fig:myu_net}, the model includes five consecutive levels in each path with the following operations: double 3D convolution (blue arrows), dropout and spatial pooling (pink arrows), upsampling (yellow arrows) and finally feature concatenation (dotted rectangles). The last layer of the model is a $1\times1$ 3D convolution (green arrow), which produces a single feature map that represents the prediction results.

\subsection{Weather Fusion UNet (WF-UNet)}
\begin{figure*}[h!]
    \centering
    \includegraphics[width=16cm]{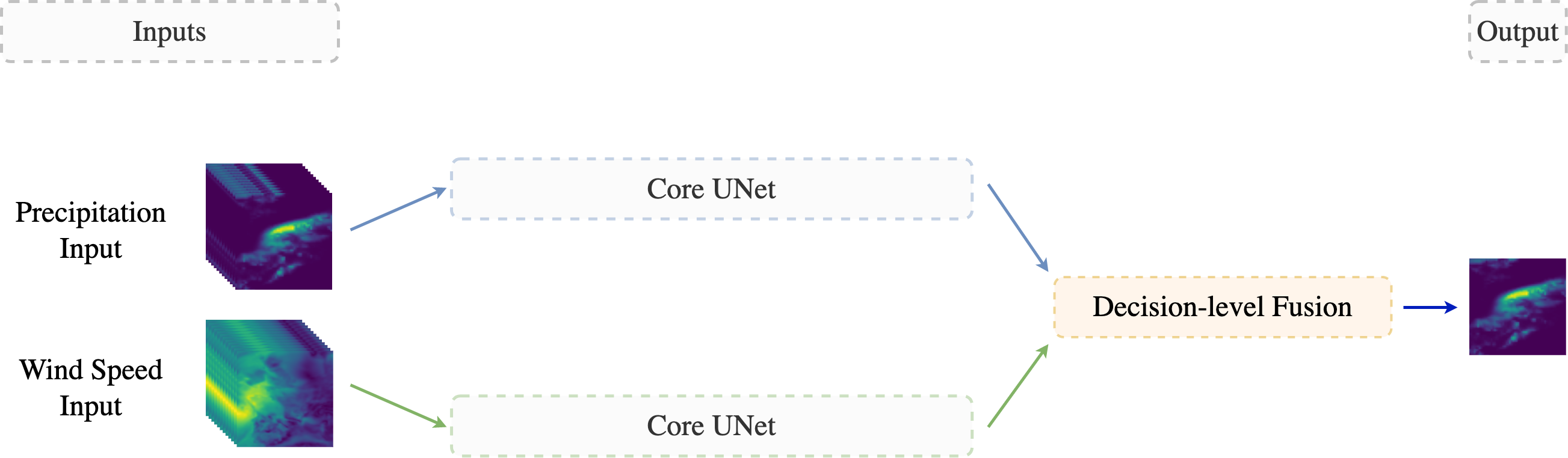}
    \caption{The architecture of WF-UNet. Two different inputs (i.e. precipition and wind speed radar images) pass through two separate but identical Core UNet streams. The outputs of those streams are then concatenated (late/decision-level fusion). Lastly, a final 3D convolution with a 1x1 kernel and a linear activation function are applied to produce the final output.}
    \label{fig:wfunet}
\end{figure*}

We propose the Weather Fusion UNet (WF-UNet) model for the precipitation nowcasting task. 
In contrast to other UNet based model which use only precipitation maps as input for precipitation nowcasting, here the proposed model utilizes past precipitation as well as wind speed radar images as input. In particular, the proposed  WF-UNet architecture, shown in Fig.  \ref{fig:wfunet}, is composed of two separate core UNet streams. Two different weather variable inputs, i.e., precipition and wind speed radar images, pass through the two streams. Here, the network processes the precipitation and wind speed images separately and then fuse the extracted features from both streams at the later decision-making stage. This architectural design is intended to address intrinsic differences between information captured in precipitation and additional radar data by encouraging each stream to learn informative features from the respective sensor separately before combining the features. The fusion is implemented via concatenating along the depth axis of the Core UNet outputs, followed by a convolutional layer. Similar to UNet, WF-UNet outputs values between 0 and 1 by applying a 3D convolution with a 1x1 kernel and a subsequent linear activation function in the final layer.

\section{Data description and preprocessing}
Here, we provide an overview of the dataset adopted for this research. 
we have selected a subset of the ERA5 
dataset \cite{https://doi.org/10.1002/qj.3803}, which includes observations of total precipitation and wind speed at the same resolution, coordinates and measurement interval.
\begin{figure}[h!]
    \centering
    \includegraphics[width=3.6cm, height=3.5cm]{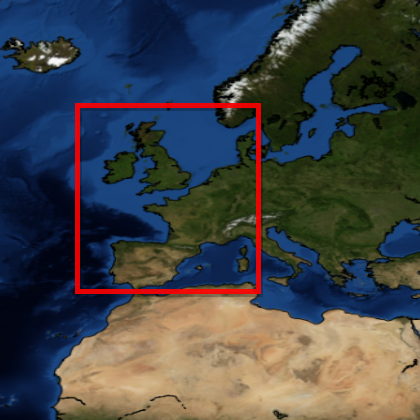}
    \caption{The study area framed in red.}
    \label{fig:study_area}
\end{figure}
The ERA5 dataset is provided by the European Centre for Medium-Range Weather Forecasts (ECMWF) through the Copernicus Climate Change Service (C3S). Through C3S, scientists combine observations of the climate system with the latest scientific findings to produce authoritative, quality-assured information about Europe's past, present, and future meteorologic conditions. 
The measurements of the selected weather variables come in grid radar images with 31 square kilometres per pixel resolution. 
Images from this region cover much of the western part of Europe within the given latitude and longitude coordinates. More specifically the study area includes all or part of 14 countries: Andorra, Belgium, Denmark, France, Germany, Ireland, Italy, Monaco, the Netherlands, Portugal, Spain, Switzerland, Luxembourg and the United Kingdom.

\begin{figure}[h!]
    \centering
    \includegraphics[width=8cm, height = 3.5cm]{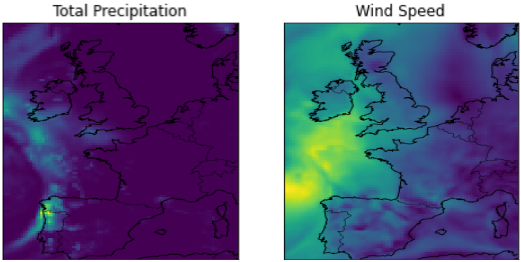}
    \caption{The dataset radar images for the different weather variable. The countries and coast outlines are included for visualization purposes only.}
    \label{fig:weather_vars}
\end{figure}

The collected dataset covers a six-year period, from 2016 to 2021, with hourly measurements. Fig. \ref{fig:study_area} and \ref{fig:weather_vars} display the geographical area we are considering and image examples of the two selected variables respectively. The \textit{Total Precipitation} is the accumulated liquid and frozen water that falls to the Earth's surface, comprising rain and snow. The values of each pixel represent the accumulated amount of rainfall in the last hour one 31 square kilometer. 
The ERA5 dataset provides the $u$ and $v$ components of wind, i.e., eastward and northward components, measured at a height of 100 metres above the surface of the Earth, in metres per second.
Given the $u$ and $v$ components, the wind speed of the wind vector $V$ is determined by $\sqrt{u^2+v^2}$. 


The collected raw radar maps have a dimension of $105\times173$ and one pixel corresponds to the accumulated rainfall or mean wind speed in the last hour on 31 square kilometers. As a data preparation step, we divided the values (separately for each weather variable) of both the training and testing set by the highest occurring value in the training set to normalize the data. Furthermore, we cropped the images to $96\times96$. 
\begin{table}[h!]
\centering
\resizebox{\columnwidth}{!}{
\begin{tabular}{c c c c c c}
\hline
\textbf{Dataset} & \textbf{Rain Pixels} & \textbf{Training} & \textbf{Validation} & \textbf{Test} & \textbf{Subset} \\ 
\hline
Original & No Extraction & 39,463 & 4,384 & 8,760 & 100\%\\ 
EU-20 & 20\% pixels & 38,906 & 4,322 & 8,677 & 98\%\\ 
EU-50 & 50\% pixels & 25,832 & 2,870 & 3,731 & 65\%\\ [1ex] 
\hline
\end{tabular}
}
\caption{Comparison of the dataset sizes. The original dataset has a lot of
images with little to no rain. (The Subset percentages are calculated based on the Training Set sizes.}
\label{table:1}
\end{table}
\begin{figure*}[h!]
    \centering
    \includegraphics[width=18cm]{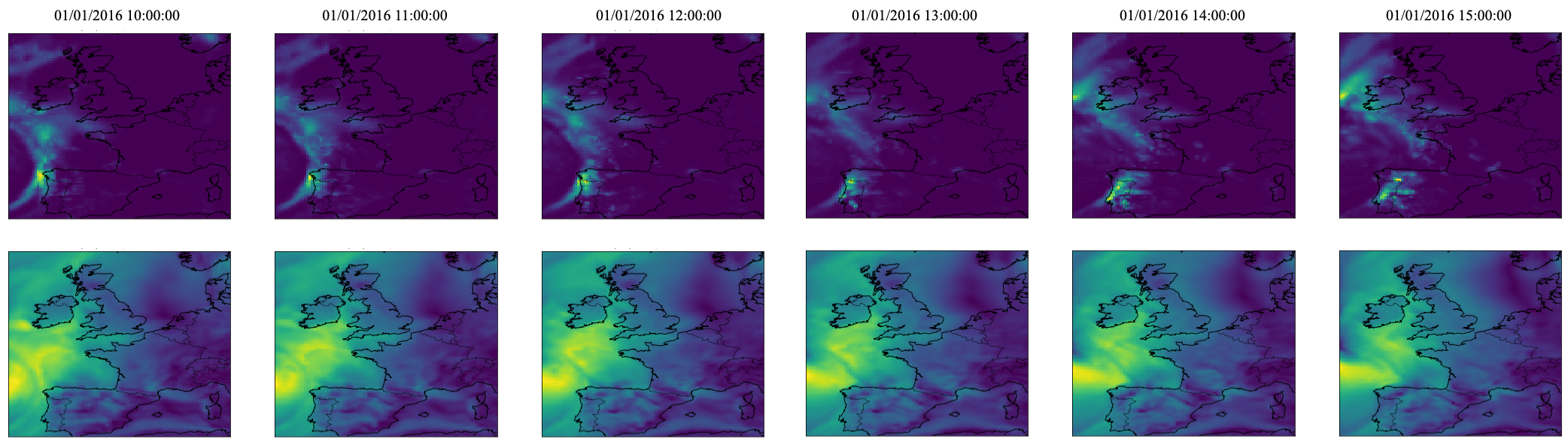}
    \caption{Example of the EU-50 dataset over 6 timesteps. The first row displays Total Precipitation radar images while the second row Wind Speed radar images.}
    \label{fig:eu50}
\end{figure*}
There is a contrast in the number of pixels showing rainfall compared to those not.  Therefore, following the lines of \cite{trebing2021smaat} and \cite{FERNANDEZ2021419}, we created two additional datasets, i.e., EU-50 and EU-20, whose images have at least 50\% and 20\% of rain in the pixels respectively.
We split the dataset into a training set (years 2016-2020) and a testing set (year 2021). Additionally, for every training iteration, a validation set is created by randomly selecting 10\% of the training set. A comparison of different sample sizes of the three datasets can be found in Table \ref{table:1}.
The EU-50 dataset is used to train and test the models. Additionally we used the test set of EU-20 to compare the generalization performance of the trained models. Fig. \ref{fig:eu50} shows an example of EU-50 dataset over 6 hours. 


\section{Experimental setup and evaluation}
The data is arranged so that the resulting inputs of total precipitation $I_{tp}$ and wind speed $I_{ws}$ are three dimensional array, i.e.,  $I_{tp}, I_{ws}\in R^{TxHxW}$. Here, $T$ is the number of lags or previous time steps corresponding to the time dimension. $H$ and $W$ refer to the size of the image and make up the spatial dimensions. TensorFlow is used to implement the models and train and evaluate them on the given dataset. 

We aim at nowcasting precipitation map for one to three hours ahead. The number of previous time-steps (lag) is set to 12, which is empirically found to be the best among the tested lag values. The height and width of the images are 96 and 96.
The model receives $I_{tp}$ and $I_{ws}$ inputs of shape $(12, 96, 96)$ and output of shape $(1, 96, 96)$. The Mean Squared Error (MSE) is used as the loss function and is optimized using Adam optimizer with an initial learning rate of 0.0001. In addition, the batch size and the dropout rate are set to 2 and 0.5, respectively. All previously described models were trained for a maximum of 200 epochs. We employed an early stopping criterion which stopped the training process when the validation loss did not increase in the last 15 epochs. 
Additionally, we used a learning rate scheduler that reduced the learning rate to a half of the previous learning rate when the validation loss did not increase for four epochs. Here, we use MSE as the primary metric to assess the model performance. Furthermore, we also include additional metrics such as accuracy, precision and recall. Following the lines of \cite{trebing2021smaat}, we first create a binarized mask of the image according to a threshold to calculate these new metrics. This threshold is the mean value of the training set from the EU-50 dataset. Hence, any value equal to or above the threshold (0.0047) is replaced by 1, and any value below it is replaced by 0. 
The models are trained on a single NVIDIA Tesla V100 with 32Gb of VRAM.

\section{Results and Discussion}
We compare the performance of our WF-UNet model with four other models, i.e., the persistence model, core UNet \cite{ronneberger2015u},  AsymmetricInceptionRes3DDR-UNet \cite{fernandez2021broad} and Broad-UNet \cite{trebing2021smaat}. The models all trained using training set of EU-50 data and tested over the test sets of both EU-50 and EU-20 datasets. 
The MSE is the main metric used for this comparison and is calculated over the denormalized data. Three additional metrics , i.e., accuracy, precision and recall are also computed over the binarized data, as described in section 5. The performance of different models on the EU-50 and EU-20 test datasets are shown in Table \ref{table:2} and \ref{table:3} respectively. 
\begin{table}[h!]
\caption{Test MSE and additional metrics values for the precipitation nowcasting obtained using the EU-50 dataset.}
\centering
\resizebox{\columnwidth}{!}{
\begin{tabular}{l c c c c c c}
\hline
\textbf{Model} & \textbf{MSE}  & \textbf{Accuracy} & \textbf{Precision} & \textbf{Recall}\\
\hline
1-hour ahead prediction\\
\hline
\textbf{Persistence} & 7.68e-03 & 0.916 & 0.803 & 0.8 \\ 
\textbf{Core UNet} & 3.18e-04 & 0.862 & 0.698 & 0.833 \\
\textbf{AsymmInceptionRes-3DDR-UNet} & 3.28e-04 & 0.859 & 0.710 & 0.781\\
\textbf{BroadU-Net} & 3.24e-04 & 0.86 & 0.705 & 0.795\\
\textbf{WF-UNet} & \underline{2.50e-04
} & \underline{0.921} & \underline{0.803} & \underline{0.849}\\
\hline
2-hour ahead prediction\\
\hline
\textbf{Persistence} & 1.56e-02 & 0.876 & \underline{0.704} & 0.697 \\ 
\textbf{Core UNet} & 5.02e-04 & 0.813 & 0.609 & 0.796 \\
\textbf{AsymmInceptionRes-3DDR-UNet} & 5.11e-04
 & 0.821 & 0.639 & 0.728\\
\textbf{BroadU-Net} & 5.05e-04 & 0.819 & 0.638 & 0.712\\
\textbf{WF-UNet} & \underline{4.62e-04
} & \underline{0.877} & 0.684 & \underline{0.813}\\
\hline
3-hour ahead prediction\\
\hline
\textbf{Persistence} & 2.19e-02 & \underline{0.861} & \underline{0.680} & 0.672 \\ 
\textbf{Core UNet} & 6.71e-04 & 0.800 & 0.612 & 0.657 \\
\textbf{AsymmInceptionRes-3DDR-UNet} & 6.45e-04
 & 0.787 & 0.583 & 0.678\\
\textbf{BroadU-Net} & 6.55e-04 & 0.806 & 0.637 & 0.609\\
\textbf{WF-UNet} & \underline{6.31e-04} & 0.855 & 0.647 & \underline{0.743}\\ \hline
\end{tabular}
}
\label{table:2}
\end{table}

\begin{table}[h!]
\caption{The obtained MSE and additional metrics values for the precipitation nowcasting on the EU-20 test dataset.}
\centering
\resizebox{\columnwidth}{!}{
\begin{tabular}{l c c c c c c}
\hline
\textbf{Model} & \textbf{MSE}  & \textbf{Accuracy} & \textbf{Precision} & \textbf{Recall}\\
\hline
1-hour ahead prediction\\
\hline
\textbf{Persistence} & 1.85e-02 & 0.932 & 0.787 & 0.787\\ 
\textbf{Core UNet} & 2.97e-04 & 0.863 & 0.698 & 0.837\\
\textbf{AsymmInceptionRes-3DDR-UNet} & 3.07e-04 & 0.861 & 0.710 & 0.786\\
\textbf{BroadU-Net} & 3.05e-04 & 0.861 & 0.706 & 0.803\\
\textbf{WF-UNet} & \underline{2.67e-04} & \underline{0.933} & \underline{0.790} & \underline{0.847}\\
\hline
2-hour ahead prediction\\
\hline
\textbf{Persistence} & 3.71e-02 & \underline{0.897} & \underline{0.680} & 0.680\\ 
\textbf{Core UNet} & 5.02e-04 & 0.813 & 0.609 & 0.796\\
\textbf{AsymmInceptionRes-3DDR-UNet} & 5.11e-04 & 0.821 & 0.639 & 0.728\\
\textbf{BroadU-Net} & 5.05e-04 & 0.819 & 0.638 & 0.712\\
\textbf{WF-UNet} & \underline{4.87e-04} & 0.895 & 0.664 & \underline{0.807}\\
\hline
3-hour ahead prediction\\
\hline
\textbf{Persistence} & 4.99e-02 & 0.874 & 0.605 & 0.606\\ 
\textbf{Core UNet} & 6.71e-04 & 0.800 & 0.612 & 0.657\\
\textbf{AsymmInceptionRes-3DDR-UNet} & 6.45e-04 & 0.787 & 0.583 & 0.678\\
\textbf{BroadU-Net} & 6.55e-04 & 0.806 & \underline{0.637} & 0.609\\
\textbf{WF-UNet} & \underline{6.34e-04} & \underline{0.877} & 0.626 & \underline{0.736}\\ \hline
\end{tabular}
}
\label{table:3}
\end{table}

\begin{figure}[h!]
    \centering
    \includegraphics[scale=0.46]{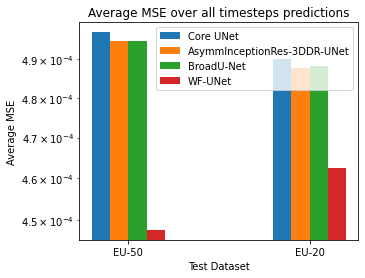}
    \caption{Average test MSE over all three timesteps ahead.} 
    \label{fig:mse_timesteps}
\end{figure}

From the obtained results, one can observe that our proposed WF-UNet model performs consistently better than all the other models for both EU-50 and EU-20 datasets and over different timestep ahead. From Table \ref{table:2}, one can notice that the largest difference in performance of WF-UNet compared to the next best performing model is for the 1-hour ahead prediction. As the number of step-ahead increases, the gap between the performance of the proposed WF-UNet and the other examined models decreases. Fig. \ref{fig:mse_timesteps}, shows the average MSE obtained using the ground truth and the nowcasts for all three timesteps ahead. One can note that the proposed WF-UNet outperforms the other tested models.

\begin{figure*}[h!]
    \centering
    \includegraphics[width=18cm, height=9cm]{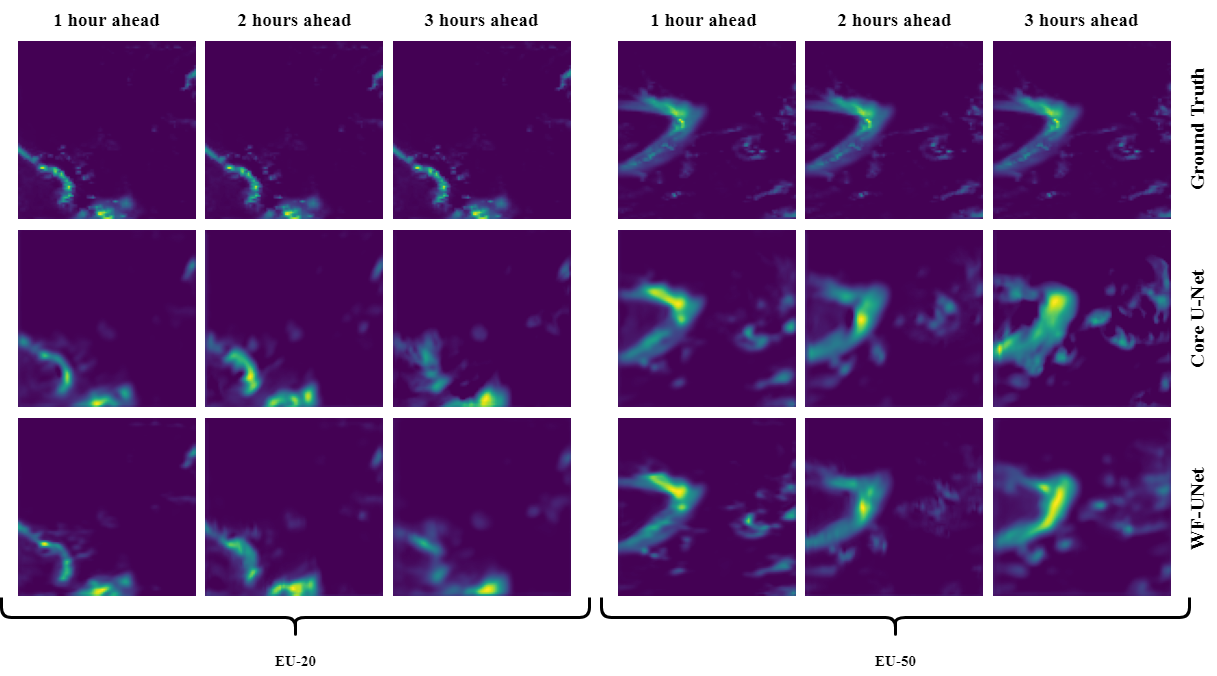}
    \caption{WF-UNet precipitation nowcasts examples. The images in the left side are generated with the test set from the dataset containing at least 20\% of rain pixels (EU-20). The images in the right side are generated with the test set from the dataset containing at least 50\% of rain pixels (EU-50).}
    \label{fig:multi_preds}
\end{figure*}

\begin{figure*}[h!]
    \centering
    \includegraphics[width=14cm]{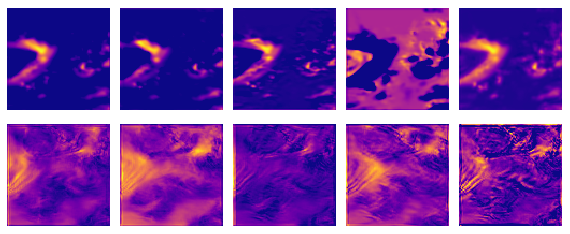}
    \caption{Feature maps outputted from the two streams before they are concatenated in the Decision-level Fusion stage. The first row represents the output of the Precipitation stream, while the second row the output of the Wind Speed stream.}
    \label{fig:feature_maps}
\end{figure*}
Two precipitation nowcasting examples are shown in Fig. \ref{fig:multi_preds}. The example on the left side is taken from the test set of the EU-20 dataset. The example on the right side is taken from the test set of the EU-50 dataset. As can be seen, the sample image of EU-50 has significantly more rain pixels than that of EU-20. 
Following the results from Tables \ref{table:2} and \ref{table:3}, the 1-hour ahead predictions in Fig. \ref{fig:multi_preds} appear to be the most similar to the ground truth image and as the hours ahead prediction increases the similarity between the nowcasts and the ground truth image decreases. Fig. \ref{fig:feature_maps} shows an example of the learned feature maps for the two streams of WF-UNet model. The image fed to the network is the ground truth image previously shown in Figure \ref{fig:multi_preds}. We can observe that different features have been extracted, some with more details than others. The obtained results show that the inclusion of multiple features such as precipitation and wind speed maps allows the WF-UNet to obtain more precise nowcasts. 
Furthermore, the decision-level fusion approach allows the network to process the two images separately and then fuse the extracted features from both streams at the later decision-making stage. With this approach, we can observe a 22\% and an 11\% improvement to the core UNet on both datasets for the 1-hour ahead predictions. The binarized predictions of the WF-UNet are 6\% more accurate than the core UNet for 1-hour and 2-hour ahead predictions and 5\% more accurate for 3-hours ahead predictions. This means that WF-UNet is better than current state-of-the-art models at predicting the position of precipitation pixels in the image. Since the model aims to perform a regression of each pixel with a wide range of values, achieving accurate forecasting or equivalently lower MSE values is more desirable than having good binary accuracy only. That is where our approach shows superior performance to the previous state-of-the-art models. 

\section{Conclusion}
The WF-UNet, an extension of the UNet architecture, is introduced for precipitation nowcasting up to three hours ahead. The model incorporates and learns from wind speed and precipitation variables using two streams. The learned features of both streams are then fused before reaching to the output of the model. The use of decision-level fusion helps to capture the spatiotemporal information of past radar images better than the classical UNet model. Compared to other tested UNet-based models, WF-UNet extracts features more efficiently, making its predictions more accurate in short-term nowcasting.
\label{sect:c}

\bibliographystyle{IEEEtran}
\bibliography{main}
\end{document}